\documentclass[letterpaper, 10 pt, conference]{ieeeconf}  % Comment this line out if you need a4paper

\IEEEoverridecommandlockouts                              % This command is only needed if 
\usepackage{hyperref}
\usepackage[table,xcdraw]{xcolor}
\usepackage{bm}
\usepackage{pgfplots}
\pgfplotsset{compat=newest} % This sets the compatibility of pgfplots
\usepackage{graphicx} % Required for inserting images
\usepackage{booktabs}
\usepackage{multirow}
\usepackage{algorithm}
\usepackage{amsmath,amssymb,amsfonts}
\usepackage{algpseudocode}
\usepackage{textcomp}
\usepackage{subfig}
\usepackage{marvosym}
\usepackage[dvipsnames]{xcolor}
\usepackage{subcaption}
\newtheorem{definition}{Definition}

\usepackage{xcolor} 

\title{\LARGE \bf
An Unsupervised Time Series Anomaly Detection Approach for Efficient Online Process Monitoring of Additive Manufacturing}

\author{
Frida Cantu\textsuperscript{*}\thanks{
This work is supported by the NSF under Grant IIS-2434916. The computational resource is supported by
Google Cloud Platform (GCP) credit. } 
\and 
Salomon Ibarra\textsuperscript{*}\thanks{* These authors contributed equally to this work.}
\and 
Arturo Gonzales~\thanks{Frida Cantu, Salomon Ibarra, Arturo Gonzales, Jesus Barreda, Li Zhang, are with the College of Engineering and Computer Science, University of Texas Rio Grande Valley, Edinburg, TX 78539 USA (email: frida.cantu02@utrgv.edu, salomon.ibarra01@utrgv.edu, li.zhang@utrgv.edu).}
\and 
Jesus Barreda
\and 
Chenang Liu\thanks{Chenang Liu is with the School of Industrial Engineering and Management, Oklahoma State University, Stillwater, OK 74078 USA (email: chenang.liu@okstate.edu).}
\and 
Li Zhang
}

\begin{document}

\maketitle
\pagestyle{empty}

%%%%%%%%%%%%%%%%%%%%%%%%%%%%%%%%%%%%%%%%%%%%%%%%%%%%%%%%%%%%%%%%%%%%%%%%%%%%%%%%
\begin{abstract}

Online sensing plays an important role in advancing modern manufacturing. The real-time sensor signals, which can be stored as high-resolution time series data, contain rich information about the operation status. One of its popular usages is online process monitoring, which can be achieved by effective anomaly detection from the sensor signals. However, most existing approaches either heavily rely on labeled data for training supervised models, or are designed to detect only extreme outliers, thus are ineffective at identifying subtle semantic off-track anomalies to capture where new regimes or unexpected routines start. To address this challenge, we propose an matrix profile-based unsupervised anomaly detection algorithm that captures fabrication cycle similarity and performs semantic segmentation to precisely identify the onset of defect anomalies in additive manufacturing. The effectiveness of the proposed method is demonstrated by the experiments on real-world sensor data.
\end{abstract}

%%%%%%%%%%%%%%%%%%%%%%%%%%%%%%%%%%%%%%%%%%%%%%%%%%%%%%%%%%%%%%%%%%%%%%%%%%%%%%%%
\section{Introduction} 

Online sensors have demonstrated their great importance in advanced manufacturing, e.g., additive manufacturing (AM)~\cite{rao2015online}\cite{shi2023lstm}\cite{nasrin2024application}. Although extensive studies in manufacturing sensor data analytics have been established~\cite{reisch2020distance}\cite{wu2022online}, how to better utilize the sensor signals for process anomaly detection is still an open research field in advanced manufacturing \cite{10.1115/MSEC2020-8503}\cite{liu2022ai}.
Recent studies in data-driven online process monitoring in manufacturing have been focusing on detecting point-based anomaly or formulating it as a distribution shift detection problem \cite{KHARITONOV20221288}. While point-based anomaly typically only focuses on extreme values, it does not work well in detecting behavior in contexts of cycle of manufacturing and some process anomalies (e.g., small shift, process alteration, etc.) might not result in an extreme value. Meanwhile, the distribution shift detection algorithms are mostly focusing on the shift of statistics such as mean and standard deviation shift, and thus they are not always suitable for applications in manufacturing. The uniqueness and challenges to developing a time series-based anomaly detection approach for a manufacturing process can be summarized as three main aspects: 1) the training data may be available in limited or no actual ground truth labels, and thereby an unsupervised learning approach is critically needed; 2) the data needs an adaptive mechanism to detect the process shifts, due to highly non-stationary nature in advanced manufacturing processes (e.g., AM processes); and 3) the application scenario expects relatively high computational efficiency, due to the practical needs for online process monitoring.

To address this challenge, this study aims to develop an unsupervised anomaly detection approach for the online sensor data in advanced manufacturing. 
We propose an unsupervised Convolutional Semantic Segmentation Anomaly Detection (CSSAD) method based on  FLUSS~\cite{gharghabi2017matrix} to accurately locate the anomaly location with a low false positive rate. In addition, we develop a convolutional optimized matrix profile computation to speed up the proposed anomaly detection approach for both CPU and GPU usage.
\begin{figure}
    \centering
    \includegraphics[width=1\linewidth]{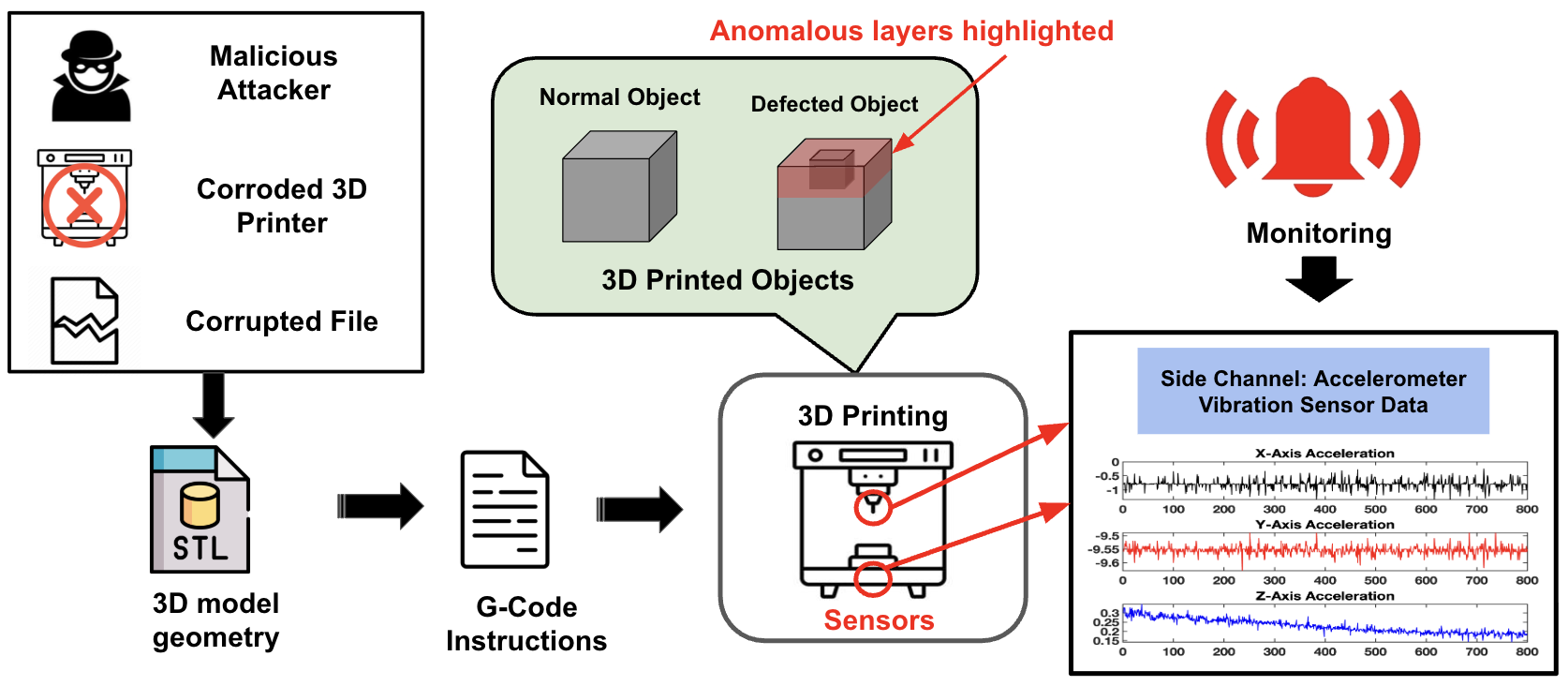}
    \caption{Problem Setting}
    \label{fig:problem-setting}
\end{figure}

\begin{figure*}[h]
  \includegraphics[width=0.95\textwidth]{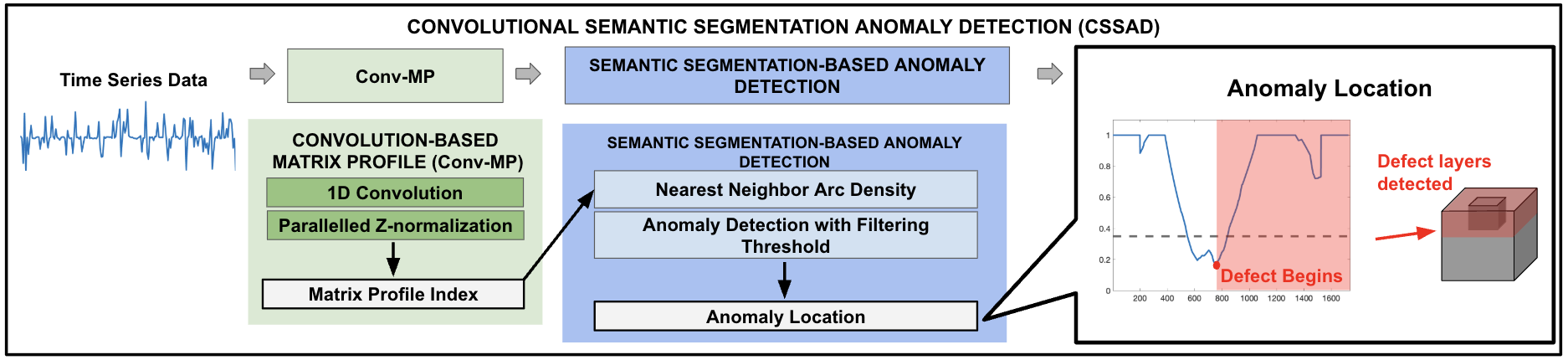}
  \caption{Overall Framework for Convolutional Semantic Segmentation Anomaly Detection (CSSAD)}
  \label{fig:framework}
\end{figure*}
\section{Problem Setting}

Figure \ref{fig:problem-setting} shows our problem setting. In 3D printing, process anomalies could come from various aspects such as a cyber physical attack or a defect. The geometric information of the 3D model is stored in the Standard Tessellation Language (STL) file, which is then translated into G-code instructions  \cite{simplify3dGcode}. 
We monitor the additive manufacturing process using data from side channels such accelerometer vibrational sensors to detect the location of anomalies during the printing process.

In unsupervised anomaly detection, there is no dependence on labels or additional information that would hint about a possible anomaly aside from what can be analyzed from the data itself. Given a time series $T$ of length $n$, the main expectation of the designed algorithm is the capability of detecting a time stamp $i$ such that the process anomaly occurs. The anomaly may cause significant alterations to the object to be manufactured. We assume before and after the occurring anomaly, the behavior of the additive manufacturing process will have some subtle differences. 

\section{Related Work}
Online monitoring is crucial for ensuring product quality and process reliability, as well as detecting cyber physical attacks. Liu et al.~\cite{LIU2022648} highlight the growing challenges of authenticating additive manufacturing (AM) data to detect potential cyber attacks due to the significant increase of AI-driven AM product development. Existing unsupervised time series anomaly detection work such as  LOF~\cite{breunig2000lof}, IsolationForest~\cite{liu2008isolation} and One-Class SVM (OCSVM)~\cite{erfani2016high} are designed for pointed based extreme values. However, in practice, many process anomalies (e.g., small shift, process alteration, etc.) might not result in an extreme value. Another line of work focus on detecting distribution shift and changing points such as TIRE~\cite{ryck2020change} utilizes a deep auto-encoder to extract time invariant features to detect if changes in the data have occurred. KL-CPT~\cite{changkernel} uses deep kernel parametrization to detect the shift. However, they are mostly designed to capture a more obvious distribution shift, and often hard to capture the subtle semantic pattern changing in printing cycle in anomalous behavior in 3D printing, and they are often less efficient compared to classical machine learning models. Recently, similarity based unsupervised anomaly detection method such as Hot-SAX~\cite{keogh2005hot}, Matrix Profile~\cite{yeh2016matrix}, Semantic Discord~\cite{zhang2020semantic} are proposed to capture the most dissimilar pattern in the entire time series. Although they identify the global anomaly sequence, the length of attack region might be unknown in practice, and could not be identified by these methods. The most closely related work is FLUSS~\cite{gharghabi2017matrix}, which is designed for semantic segmentation problem in time series. However, they are not designed for this problem and could have high false positive rate.

\section{Background}

In this section, we will explain definitions and preliminaries for time series we will use in this paper. 
\begin{definition}
\textbf{Time series}: Time series $T$ is an ordered real numbers of length $n$. \text{$T = [t_1, t_2,..., t_n]$}. 
\end{definition}
\begin{definition}
\textbf{Subsequence}: A subsequence $T_{i,m}$ is defined as a continuous set of data points of length $m$ in $T$ that starts at the index $i$, where $1 < i \leq n-m+1$. 
\end{definition}

In the additive manufacturing process, accelerometer sensor data records rich information of each printing layer, which are typically similar across layers. With a user-defined subsequence length, and appropriate distance metric, it is easy to track and monitor the behavior of printing quality. In time series data mining, we often use Z-normalized Euclidean distance to reset the offset and the scale ~\cite{keogh2002need}\cite{zhang2020semantic}.

\begin{definition}
\textbf{Z-Normalized Euclidean distance} refers to applying Z-score normalization to our Euclidean distances $D(i,j)$ between subsequences $T_{i,m}$ and $T_{j,m}$. The equation is formally described below:
\begin{equation}
D_z(i,j) = \sqrt{ \sum_{p=1}^{m} 
\left( 
\frac{t_{i - p + 1} - \mu_{i,m}}{\sigma_{i,m}} 
- 
\frac{t_{j - p + 1} - \mu_{j,m}}{\sigma_{j,m}} 
\right)^2 }
\label{eq:z-normED}
\end{equation}

where $\mu_{p,m}$ and $\sigma_{p,m}$ are the mean and standard deviation of subsequence $T_{p,m}$, and $\mu_{q,m}$ and $\sigma_{q,m}$ are the mean and standard deviation of subsequence $T_{q,m}$, respectively. \end{definition}

Z-normalization ensures the comparison is well aligned, so we could align the shape of the printing layer cycles in 3D printing process. 
\begin{definition}
    \textbf{Nearest Neighbor Subsequence}: The nearest neighbor of subsequence $T_{i, m}$ in time series $T$ is the subsequence $T_{j, m}$  of length $m$ with the smallest Z-normalized Euclidean distance with subsequence $T_{i, m}$. 
\end{definition}
The nearest neighbor of a subsequence of a production pattern would identify a similar production pattern in the entire manufacture process. One could simply store the location of the nearest neighbor's distance and its z-normalized distance, which is known as \textbf{Matrix Profile} \cite{yeh2016matrix}. 

\begin{definition}
    \textbf{Matrix Profile} is a $2 \times (n-m+1)$ matrix that holds the nearest neighbor index and its nearest neighbor distance for every subsequence  $T_{m, i}$ in $T$. 
\end{definition} 

\section{Proposed Method}
Figure \ref{fig:framework} shows our overall framework. Our proposed method consists of two components: Convolution-based Matrix Profile (Conv-MP) and Semantic Segmentation-based anomaly detection algorithm. We first compute Convolution-based Matrix Profile, then our second component computes nearest neighbor arc density curve through semantic segmentation using bias correction. Finally, we design a threshold parameter to identify the anomaly location. 
\vspace{-2mm}

\subsection{Background on STOMP for Matrix Profile}
The current state-of-the-art STOMP~\cite{zhu2016matrix} proposes an efficient way to compute matrix profile by caching distance computation and compute in sequentially. STOMP first computes a sliding dot product vector $QT_{i,n-m+1}$ by caching the previous computation $QT_{i-1, j-1}$ diagonally as 
\begin{equation}
\label{eqn:QT_seq}
    QT_{i,j} = QT_{i-1,j-1} - t_{i-1}t_{j-1}+ t_{i+m-1}t_{j+m-1}
\end{equation}
where $QT_{i,j} = \sum_{l=1}^{m} T_{i,m}T_{j,m}$, which is dot product of the sequence $T_{i, m}$ and $T_{j, m}$. Then they compute the Z-normalized $D_Z(i, j)$ parallel as:
\begin{equation}
    D_z(i,j) = \sqrt{2m(1 - \frac{QT_{i,j}- m\mu_i\mu_j}{m\sigma_i\sigma_j}}),  
    \label{eq:Z-normED2}
\end{equation}
Where $m$ is the subsequence length, $QT_{i,j}$ is the dot product of subsequences $T_{i.m}$ and $T_{j,m}$, and $\mu_i$ and $\mu_j$ are the means and standard deviations of subsequences $T_{i.m}$ and $T_{j,m}$ respectively. However, it still needs to compute $QT_{i, j}$ sequentially and does not utilize the parallel computational ability in devices like GPU. 

\subsection{Convolution-based Matrix Profile} 

To address the above challenge and allow modern GPU parallel computing such as Pytorch, we propose Convolution-based Matrix Profile (Conv-MP). Particularly, Conv-MP consists of two modules: convolutional dot product and parallel z-normalization.\\ 
\noindent\textbf{1D convolution} We use 1D convolution to Given an input kernel $Q$ and a time series $T$, 1D convolution $Q*T$ is defined as
\begin{equation}
    Q*T = \sum_{j=1}^{n-m+1} Q\cdot T_{j, m}
\end{equation}
where $Q$ is a kernel of size $1\times m$, and $*$ denote the dot product. 
We utilize the property of dot product could be computed in parallel on GPU, that means if $Q'$ is multiple kernels of size of $k\times m$, it would take the same time for a GPU to compute $Q'*T$ and $Q*T$. 

\noindent \textbf{Parallel Z-normalization} Instead of computing Z-normalization sequentially, we apply parallel Z-normalization on the dot product inspired by Equation \ref{eq:Z-normED2}. Let $S$ denote the set of all sliding windows (subsequences) of length $m$ extracted from the time series $T$, and $\bm{\mu}$ and $\bm{\sigma}$ denote vectors containing the moving means and standard deviations of all subsequences in $S$. We can apply parallel z-normalization on top of the outcome of 1D convolution $S*T$. Specifically, we can fast compute the query of subsequence $T_{i, m}$ to time series $T$ as follows:   
\begin{equation}
   D_z(i,:) = \sqrt{2m(1-\frac{S*T-m\mu_i \bm{\mu}}{m\sigma_i \bm{\sigma}}).}
    \label{eq:Z-normED3}
\end{equation}

\begin{algorithm}[!t]
\small
\caption{Convolution-based Matrix Profile (convMP)}
\label{algo1}
\begin{algorithmic}[1]
    \State \textbf{input}: {Time series $T$ of length $n$ subsequence length $m$}
    \State \textbf{output}: Matrix profile $P$ of length $n-m+1$ and associated matrix profile index $I$ of length $n-m+1$
   \State $l = length(T)-m+1$ {\color{blue}// length of $P$ and $I$}
   \State $eZone = m/4$ {\color{blue}//initializing our exclusion zone}
   \State {\color{blue}// Extract all the subsequences of $T$ of length $m$}
   \State $S = slidingWindow(T, m, 1)$
   \State {\color{blue}// calculating the means and std of all subsequences}
   \State $\sigma, \mu = calcMeanStd(S)$ 
   \State {\color{blue}// Initializing variables of $1\times l$ dimensions}
   \State $P,I = empty(l)$
    \State {\color{blue}// Apply convolution to $T$ with $S(i)$}
    \State $QT = 1D\text{-}Convoluiton(T, S)$
    \State {\color{blue}// Calculate our i-th distance profile}
    \State $D = calculateDist(QT, \sigma, \mu, m)$
    {\color{blue}// See Eqn~\ref{eq:Z-normED3}}
    \State{\color{blue}// Mask self-matching values index within $\pm eZone$.}
    \State $applyMask(D,eZone$)
    \State $P = min(D), I = argmin(D)$
    \State \textbf{return} $P,I$
\end{algorithmic}
\end{algorithm}

Algorithm \ref{algo1} shows our proposed convolution-based Matrix Profile (ConvMP). Line 3-4 initialize the variables. Line 6 computes the moving mean and standard deviation vectors of the time series. Line 8 extracts all subsequences in $T$ as $S$ a $m\times l$ matrix. $S$ will be used as our kernel for 1D Convolutional network. Line 11-19 computes the Matrix Profile by first computing the dot product vector $QT_i$  for each subsequence through convolution 1D, then computing the distance through Eqn (\ref{eq:Z-normED3}). Since convolution and parallel computing are utilized, we improved the runtime for computing Matrix Profile on GPU.

\subsection{Semantic Segmentation-based Anomaly Detection}

Our semantic segmentation-based anomaly detection algorithm. Following our problem setting, we assume that the anomaly routine would inherently change into a different behavior with existing printing at the beginning stage, inspired by FLUSS~\cite{gharghabi2017matrix}, we first perform semantic segmentation followed by anomaly detection with post threshold filtering to reduce false positives and report the anomalous layer detected.

\begin{figure}
    \centering
    \includegraphics[width=1\linewidth]{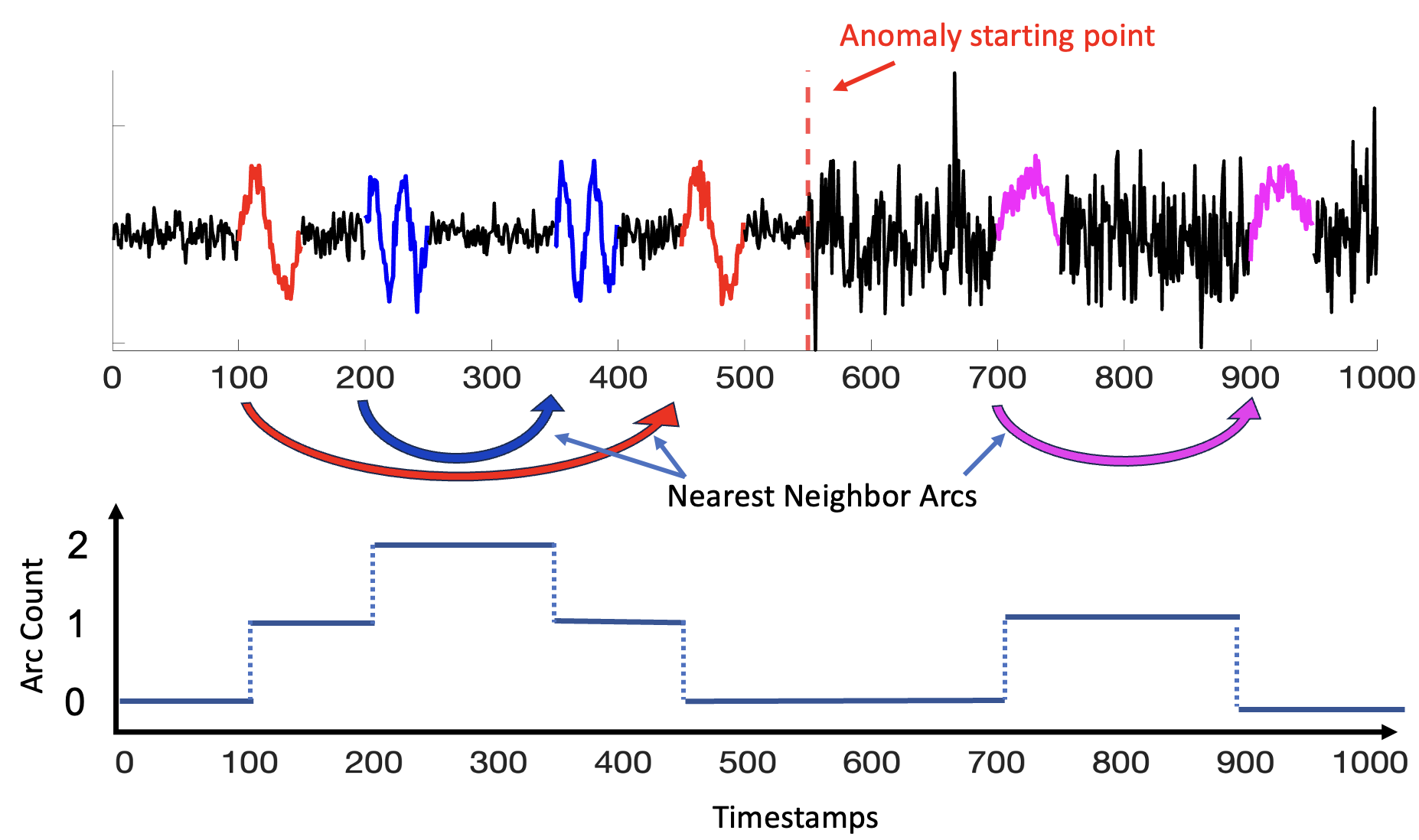}
    \caption{top: nearest neighbor arc formed by nearest neighbor subsequence pairs $T_{100, 50}$ and $T_{450, 50}$ (in red). Bottom: arc count over time series $T$.  }%Illustration of nearest neighbor arcs ( top) and the arc count corresponding to time stamp in a time series (bottom).}
    \label{fig:arc_example}
\end{figure}

\noindent\textbf{Nearest Neighbor Arc Density} We follow~\cite{gharghabi2017matrix} to compute Nearest Neighbor Arc Curve and Arc Count. Particularly, given Matrix Profile Index of time series $T$, we first form arcs between every subsequence $T_{i, m}$ and its nearest neighbor subsequences $T_{j, m}$, and then compute the number of neighbor arc counts crossing over each time step $t$ for time series. We use $Arc(i, j) $ to denote the nearest neighbor arc formed by nearest neighbor pairs $T_{i, m}$ and $T_{j, m}$.  Figure \ref{fig:arc_example} (top) shows three nearest neighbor arc formed by three pairs of $Arc(100, 450)$ (in red), $Arc(200, 350)$ (in blue) and $Arc(700, 900)$ (in purple). We could then obtain the Arc Count (AC) going over each timestamp in the time series as shown in Figure \ref{fig:arc_example}.bottom. We then apply a bias correction and a normalization to the original arc count to account for the tendency of more arc curves to intersect the center, where fewer curves are typically represented. Specifically, corrected Arc Density $CAD_i$ for a subsequence $T_{i, m}$ in time series $T$ could be computed as: 
\begin{equation}
     AC_i = min\left(\frac{AC_i}{IAC_i},1\right)
     \quad 
    CAD_i = \frac{AC_i}{\max(AC)}.
  \label{eqn:cac}
\end{equation} 
where the theoretically random arc density can be computed as $IAC_i = \frac{2i (l - i)}{l}\label{eq:acnorm}$ from \cite{gharghabi2017matrix}.

\textbf{Anomaly Detection with Filtering Threshold $\epsilon$}
Since nearest neighbors tend to stay within its own similar region rather than cross over to another zone, with an appropriate filter strategy, we can use this idea to detect changing in printing behavior and identify anomalous layer. We find the minimum value in corrected arc density (CAD) as the start of anomaly location. 
Given a time series $T$ and corrected arc density $CAD$ vector of length $n$, if we use 1 to denote anomaly and 0 to denote normal, then the anomaly output $y_i$ for subsequence $T_{i, m}$, 
 \[ y_i = \begin{cases} 
      0 & i\leq \arg min(CAD) \\
      1 & i> \arg min(CAD) 
   \end{cases}\]

\begin{figure}[h]
    \centering
\includegraphics[width=\linewidth]{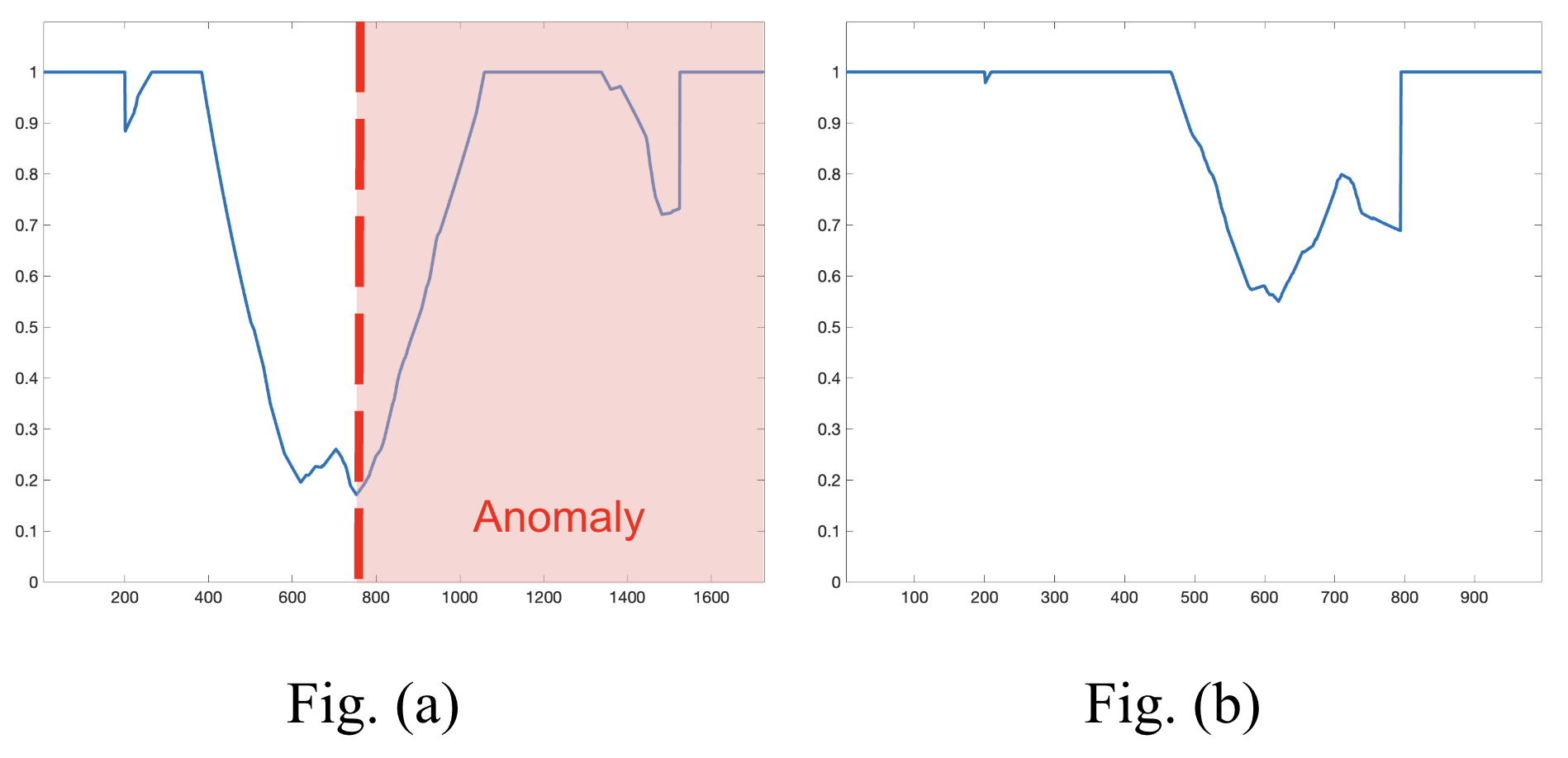}
\caption{(a) Corrected density plots
(CAD) of a sensor data containing anomalous defects
(i.e. anomaly data) with ground truth defects (anomaly region) highlighted. (b) CAD of on the same data removing the defects part (i.e., clean data). }
\label{fig:caseA}
\end{figure}

Figure~\ref{fig:caseA} (a) and (b) shows the corrected density plots $CAD$ between a sensor data containing anomalous defects (aka anomaly data) and the data removing the defects part (aka. clean data). We observe that data with anomaly has significant lower value at $CAD$ curve due to more obvious defects, whereas the clean data has less significant fluctuations during printing. 

Based on this observation, we propose to use a threshold filter technique to reduce false alarms due to normal fluctuation in printing shapes. Specifically, given a sensitivity threshold $\epsilon$, we only report an anomaly if $\min(CAD)$ is lower than $\epsilon$ to ensure this is a significant change, otherwise we do not report any anomaly for time series $T$. In our experiment, we will conduct a parameter test to show the effect in $\epsilon$ and discuss the impact on anomaly detection performance. 

\section{Experiments} 
We conducted experiments to show the effectiveness of our proposed anomaly detection algorithm against other baselines, an ablation study, and a parameter test on the threshold.

\begin{figure}
    \centering
    \includegraphics[width=0.8\linewidth]{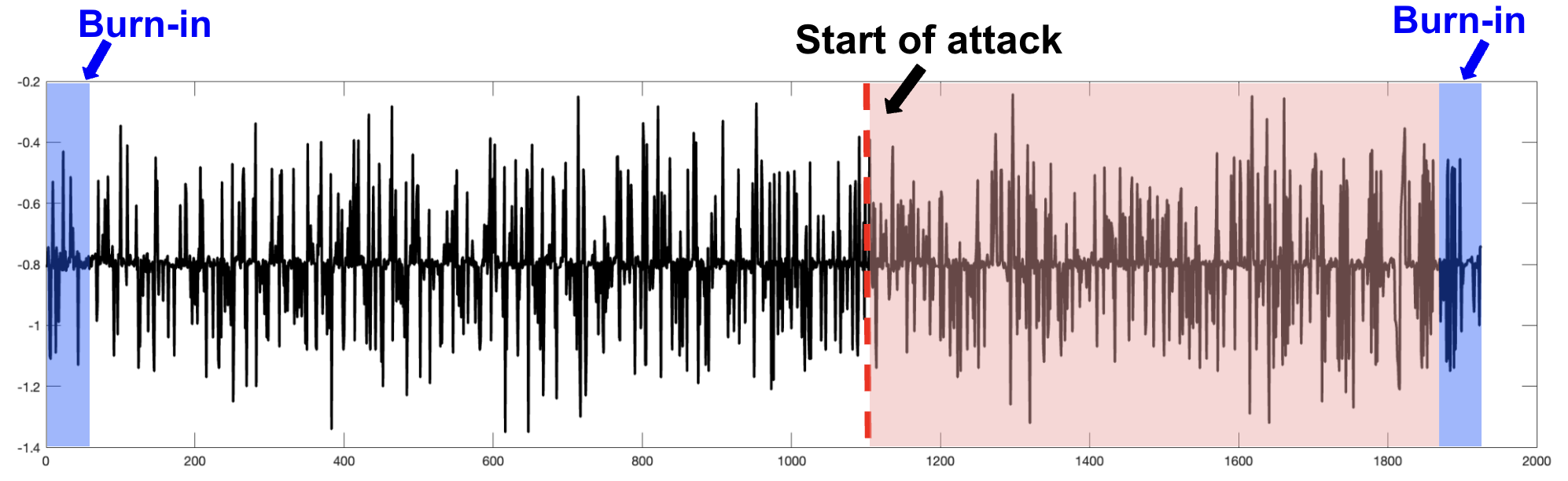}
    \caption{The x-axis on one of the 3D printer sensor datasets with red highlighting the area affected by the attack and blue highlighting the burn-in part of the data that was excluded.}
\label{fig:x-axis-data}
\end{figure}
\vspace{-1mm}
\subsection{Datasets}
\vspace{-1mm}
For the experiments, we evaluated the effectiveness of our method on 7 real-world 3D printing sensor data from Zhou et al. \cite{zhouCASE2021}. Each sensor data is ranging 1700-1900 data points  collected over 30 minutes printing from micro-electro-mechanical system (MEMS) accelerometers mounted on a fused filament fabrication (FFF)-based 3D printer coming from \cite{Shi2022LSTM}. As mentioned by Shi et al. 2022 \cite{Shi2022LSTM}, sensors were attached to both the extruder and the printing bed, recording real-time vibrations along the x, y, and z axes at a sampling frequency of approximately 3 Hz, at a printing speed of 40 mm/s for 67 total layers \cite{zhouCASE2021}. The data captured the printing process of an approximately 2 cm cube, which was subjected to a cyber-physical attack. For the experiments, we determined that the x-axis of each dataset best represented the impact of the cyber-physical on the data compared to the other axes due to the high probability of false positives the x-axis showed compared to the other axes. In figure \ref{fig:x-axis-data}, we plotted out the x-axis of one of our datasets and marked using a red dashed line the ground truth for where the anomaly is located, which is after the first intital $60\%$ of the dataset, and affected the behavior of the 3D sensor time series.

\iffalse 
\begin{table}[]
\caption{\fontseries{m}\selectfont Real-world datasets length including burn-in, without burn-in, and length of clean section in the datasets}
\centering
\scalebox{0.9}{
\begin{tabular}{@{}c c c c@{}}
\toprule
Datasets   & Length (Burn-in) &  Length (no Burn-in) & Clean Length \\
\hline
1  & 1926 & 1866 & 1126     \\
2  & 1886 & 1826 & 1102  \\
3  & 1883 & 1823 & 1100 \\
4  & 1884 & 1824 & 1100 \\
5  & 1885 & 1825 & 1101 \\
6  & 1885 & 1825 & 1101 \\
7  & 1909 & 1849 & 1115 \\
\hline
Average  & 1894 & 1834 & 1106  \\
\bottomrule
\end{tabular}}
\label{tab:data_size}
\end{table}
\fi 

\begin{table*}[]
\caption{\fontseries{m}\selectfont Overall Performance (F1-score and False Positive Rate and the total running time comparison in CPU and GPU. (-) means not specially designed for GPU. }
\centering
\scalebox{0.97}{
\begin{tabular}{ccccccccccccr}
\hline
\multicolumn{1}{l}{}                         & \multicolumn{2}{c}{Proposed} & \multicolumn{2}{c}{LOF}   & \multicolumn{2}{c}{OCSVM} & \multicolumn{2}{c}{IF}    & \multicolumn{2}{c}{TIRE}  & \multicolumn{2}{c}{KL-CPD}      \\ \cline{2-13} 
\multicolumn{1}{l}{Dataset Index}                  & F1 $(\uparrow)$                & FPR $(\downarrow)$     & F1 $(\uparrow)$         & FPR $(\downarrow)$         & F1 $(\uparrow)$         & FPR   $(\downarrow)$       & F1 $(\uparrow)$          & FPR $(\downarrow)$        & F1 $(\uparrow)$           & FPR $(\downarrow)$     & F1 $(\uparrow)$    & \multicolumn{1}{c}{FPR} $(\downarrow)$ \\ \hline
1                                            & \textbf{0.835}     & 0.264     & 0.284       & 0.058       & 0.428       & 0.506       & 0.153        & 0.326      & 0.488        & 0.020      & 0.207 & 0.333                   \\
2                                            & \textbf{0.831}     & 0.159     & 0.287       & 0.078       & 0.416       & 0.507       & 0.147        & 0.330      & 0.486        & 0.021      & 0.207 & 0.332                   \\
3                                            & \textbf{0.811}     & 0.200     & 0.286       & 0.068       & 0.396       & 0.516       & 0.152        & 0.328      & 0.489        & 0.020      & 0.207 & 0.334                   \\
4                                            & \textbf{0.801}     & 0.220     & 0.318       & 0.125       & 0.422       & 0.498       & 0.154        & 0.326      & 0.490        & 0.016      & 0.206 & 0.334                   \\
5                                            & \textbf{0.820}     & 0.179     & 0.337       & 0.118       & 0.428       & 0.499       & 0.142        & 0.333      & 0.491        & 0.017      & 0.206 & 0.333                   \\
6                                            & \textbf{0.753}     & 0.329     & 0.281       & 0.092       & 0.421       & 0.498       & 0.150        & 0.328      & 0.489        & 0.021      & 0.206 & 0.334                   \\
7                                            & \textbf{0.831}     & 0.160     & 0.290       & 0.059       & 0.429       & 0.486       & 0.155        & 0.326      & 0.485        & 0.021      & 0.207 & 0.333                   \\ \hline
\multicolumn{1}{l}{Average}                  & \textbf{0.812}     & 0.200     & 0.298       & 0.085       & 0.420       & 0.501       & 0.150        & 0.328      & 0.488        & 0.019      & 0.207 & 0.333                   \\ \hline
\multicolumn{1}{l}{Run Time(CPU)(s)} & \multicolumn{2}{c}{1.118}      & \multicolumn{2}{c}{0.133} & \multicolumn{2}{c}{0.992} & \multicolumn{2}{c}{1.129} & \multicolumn{2}{c}{3.937} & \multicolumn{2}{c}{10800+}      \\
\multicolumn{1}{l}{Run Time(GPU)(s)} & \multicolumn{2}{c}{0.910}      & \multicolumn{2}{c}{-} & \multicolumn{2}{c}{-} & \multicolumn{2}{c}{-}  & \multicolumn{2}{c}{4.145} & \multicolumn{2}{c}{2012.077}    \\ \hline
\end{tabular}}
\label{tab:baselinecomparison}
\end{table*}

\subsection{Experiment setup} 
We test the F1 and performance, running time and false alarm rate in real world data.
\begin{itemize}
    \item How good is our proposed method identifying cyber physical attacks compared to existing work in real world addictive manufacture process? 
    \item How to choose threshold parameter $\epsilon$ to reduce false alarms in practice? 
\end{itemize}
To answer the first question, we conduct our first experiment on seven datasets where we evaluate the performance of our proposed method in comparison to a few anomaly detection based baselines. The second experiment will focus on testing our method and measuring the false positive rate on the initial $60\%$ of each dataset, which is the clean datasets made from the section unaffected by the anomaly, and how the addition of a threshold on the anomaly detection algorithm affects the false positive rate. We are using the subsequence length of $100$ and an arbitrary threshold $\epsilon$ of 0.35. The data was also shorten a bit by excluding the first and last 30 data points to account for 3D printing burn-in.
\iffalse
\begin{table}[]
\caption{\fontseries{m}\selectfont False Positive Rate Performance on 7 Derived Datasets from the 3D Printing Sensor Data}
\centering
\scalebox{1.1}{
\begin{tabular}{@{}c c c c@{}}
\toprule
Clean \\ Datasets   & w/o threshold & with threshold & Improvement \\
\hline
1  & 0.330 & 0.938 & +0.608      \\
2  & 0.083 & 0.937 & +0.854      \\
3  & 0.344 & 0.937 & +0.593      \\
4  & 0.285 & 0.285 & +0.000      \\
5  & 0.350 & 0.937 & +0.587      \\
6  & 0.347 & 0.937 & +0.590      \\
7  & 0.875 & 0.938 & +0.063      \\
\hline
Average  & 0.373 & 0.844 & +0.471   \\
\bottomrule
\end{tabular}}
\label{tab:ablation}
\end{table}
\fi

\subsection{Baseline Methods}
We will compare our method with the following baseline methods in our experiments. \textbf{Local Outlier Factor (LOF)} proposed by Breunig et al.~\cite{breunig2000lof}. LOF detects points with lower local density compared to others in our data and points those out as anomalies. \textbf{IsolationForest (IF)} proposed by Liu et al.~\cite{liu2008isolation}. IF isolates anomalies by partitioning the data, anomalies become apparent due to their distinct patterns.\textbf{One-Class SVM (OCSVM)} proposed by Erfani et al.~\cite{erfani2016high}. OCSVM learns the boundary of normal data in a high-dimensional feature space and labels anything that falls outside the boundary as an anomaly. \textbf{Time-Invariant Representation (TIRE)} proposed by Ryck et al. ~\cite{ryck2020change}. TIRE uses deep auto encoders to detect invariant features, and the reconstruction loss to detect anomalies. \textbf{Kernel Learning Point Detection (KL-CPD)} proposed by Chang et al.\cite{chang2019kernel},
uses deep kernel parametrization to detect distribution shift in our data. In our experiment, we used a subsequence length of 100 for our method. For the Isolation Forest (IF) baseline, we set the maximum samples parameter to 20, and for One-Class SVM (OCSVM), we set gamma to `auto'. All other baselines and parameters were kept as default. The experiments were conducted using an NVIDIA L4 GPU on the Google Cloud Platform.
\vspace{-2mm} 

\subsection{Evaluation Metrics}

 %In formula~\ref{eq:f1}, precision and recall make up the F1 formula. 
 We use F1-score to measure the performance of anomaly detection and False Positive Rate (FPR) to measure the false alarm rate. The equation of F1-score and FPR are:  
\begin{equation*}\label{eq:f1}
\vspace{-2mm}
F1 = \frac{2\cdot \text{precision}\cdot\text{recall}}{ \text{precision} + \text{recall}} \quad\quad  FPR = \frac{\text{FP}}{\text{FP} + \text{TN}}
\end{equation*}
\noindent where the $\text{precision} = \frac{TP}{TP+FP}$
$\text{recall} = \frac{TP}{TP+FN}$. 
F1 score is the geometric mean of precision and recall. Precision is a measure of how good a method is able to report true positives in the total predicted positives, and recall measures the quantity of false negatives that were missed.

\subsection{Experiment 1: Performance on Anomaly Detection in Real-world data. }

Table~\ref{tab:baselinecomparison}, shows our result F1 score, FPR, and running time for all seven datasets against five anomaly detention baselines. Our proposed anomaly detection method obtained an average F1 score of 0.812, greatly exceeding all other baselines (the second best is TIRE with 0.488). We have the second-best false positive rate of 0.2. in CPU we are only significantly slower than LOF, and slightly lose to OCVSM with a difference of 0.126 seconds. However, with our GPU acceleration with convolution, we improve 18.4\% improvement from our own CPU time, Overall, we are the best in all GPU methods and second best among all methods running in CPU and GPU. 

\vspace{-2mm}
\subsection{Experiment 2: Ablation Study on Threshold Parameter}
\begin{table}[ht]
\centering
\caption{False Positive Rate (FPR) Comparison With and Without \(\epsilon\)}
\label{tab:fpr_epsilon}
\begin{tabular}{c cc}
\hline
Dataset Index & \multicolumn{2}{c}{FPR} \\
\cline{2-3}
 & Without $\epsilon$ & With $\epsilon$ \\
\hline
1 & 0.428 & \textbf{0.000} \\
2 & 0.465 & \textbf{0.000} \\
3 & 0.190 & \textbf{0.000} \\
4 & 0.384 & 0.384 \\
5 & 0.473 & 0.473 \\
6 & 0.448 & 0.448 \\
7 & 0.248 & \textbf{0.000} \\
\hline
\end{tabular}
\label{tab:cleandatathreshold}
\end{table}

In this experiment, we will perform an ablation study on the threshold parameter to test on how $\epsilon$ impacts false positive rate when using clean anomaly-free data as input.
In Table \ref{tab:cleandatathreshold}, we show the false positive rate of the clean data without our threshold, $\epsilon$, implementation. The last column in the table show the false positive rate results after applying our algorithm with $\epsilon$ to the clean data. We bold the results that demonstrate how the addition of using $\epsilon$ made the false positive rate decrease substantially. The results that stayed the same even after the addition of $\epsilon$ tell us that some of the clean data's minimum $CAD_i$ were lower than the $\epsilon$ of 0.35, which might indicate the need for the user to adjust the $\epsilon$ to be lower. This experiment showed that the $\epsilon$ was effective in reducing the false positive rate.

\subsection{Experiment 3: Parameter test}

\begin{figure}[h]
    \centering
    \includegraphics[width=0.6\linewidth]{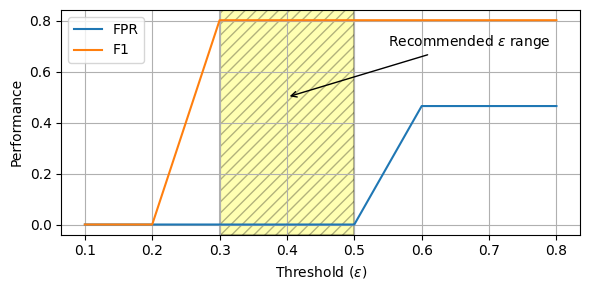}
    \caption{results of our parameter test} 
    \label{fig:rec}
\end{figure}

In this experiment, we conduct a sensitivity analysis to test an effective range of the hyperparameter $(\epsilon)$ to test on the impact on F1 score and False positive rate (FPR). For each value, we report the F1 score on dataset 2 in our experiment 1, as well as FPR on its anomaly free version. Figure \ref{fig:rec} shows our performance of F1-score (in orange) and FPR (in blue) with $\epsilon$ varies from .1 to .8 in increments of 0.1. Our results show that a high threshold (greater than 0.6) will result in a high F1 score but significantly increase in false alarm, where as an overly low threshold (lower than 0.2) might maintain a low false positive but significantly reduce F1 score. From our data, we recommend selecting $\epsilon$ within the range $0.3 \le \epsilon \le 0.5$. In practice, if anomaly-free data is available, one could set the threshold to the minimum density curve value to balance the trade-off between F1 and FPR. 

\section{Conclusion}
In this work, we propose an unsupervised anomaly detection based on semantic segmentation in high-resolution In this work, we propose an unsupervised anomaly detection based on semantic segmentation in high resolution sensor time series. Our proposed method is able to perform online monitoring by accurately and efficiently locating the anomaly location with a low false positive rate. In addition, we also propose a convolution-based distance computation module to allow convolutional operation on potential available GPU devices. Our experiment shows our proposed method achieves a higher F1 score and lower false positive than existing method with a better efficiency in both CPU and GPU. 

In the future, we would like to extend our current framework to develop an online batch-version algorithm to incrementally update the anomaly location and capture potential shift with more complex printing objects. 
\addtolength{\textheight}{-12cm}   % This command serves to balance the column lengths

%%%%%%%%%%%%%%%%%%%%%%%%%%%%%%%%%%%%%%%%%%%%%%%%%%%%%%%%%%%%%%%%%%%%%%%%%%%%%%%%

%%%%%%%%%%%%%%%%%%%%%%%%%%%%%%%%%%%%%%%%%%%%%%%%%%%%%%%%%%%%%%%%%%%%%%%%%%%%%%%%

%%%%%%%%%%%%%%%%%%%%%%%%%%%%%%%%%%%%%%%%%%%%%%%%%%%%%%%%%%%%%%%%%%%%%%%%%%%%%%%%
%\section*{APPENDIX}


%\section*{ACKNOWLEDGMENT}
%This work is supported by the NSF under Grant IIS-2434916. The computational resource is supported by Google Cloud Platform (GCP) credit.

\bibliographystyle{abbrv}
\bibliography{attack}

\begin{thebibliography}{10}

\bibitem{10.1115/MSEC2020-8503}
{\em An Online Side Channel Monitoring Approach for Cyber-Physical Attack Detection of Additive Manufacturing}, volume Volume 2: Manufacturing Processes; Manufacturing Systems; Nano/Micro/Meso Manufacturing; Quality and Reliability of {\em International Manufacturing Science and Engineering Conference}, 09 2020.

\bibitem{breunig2000lof}
M.~M. Breunig, H.-P. Kriegel, R.~T. Ng, and J.~Sander.
\newblock Lof: identifying density-based local outliers.
\newblock In {\em Proceedings of the 2000 ACM SIGMOD international conference on Management of data}, pages 93--104, 2000.

\bibitem{changkernel}
W.-C. Chang, C.-L. Li, Y.~Yang, and B.~P{\'o}czos.
\newblock Kernel change-point detection with auxiliary deep generative models.
\newblock In {\em International Conference on Learning Representations}.

\bibitem{chang2019kernel}
W.-C. Chang, C.-L. Li, Y.~Yang, and B.~P{\'o}czos.
\newblock Kernel change-point detection with auxiliary deep generative models.
\newblock {\em arXiv preprint arXiv:1901.06077}, 2019.

\bibitem{erfani2016high}
S.~M. Erfani, S.~Rajasegarar, S.~Karunasekera, and C.~Leckie.
\newblock High-dimensional and large-scale anomaly detection using a linear one-class svm with deep learning.
\newblock {\em Pattern Recognition}, 58:121--134, 2016.

\bibitem{gharghabi2017matrix}
S.~Gharghabi, Y.~Ding, C.-C.~M. Yeh, K.~Kamgar, L.~Ulanova, and E.~Keogh.
\newblock Matrix profile viii: domain agnostic online semantic segmentation at superhuman performance levels.
\newblock In {\em 2017 IEEE international conference on data mining (ICDM)}, pages 117--126. IEEE, 2017.

\bibitem{keogh2002need}
E.~Keogh and S.~Kasetty.
\newblock On the need for time series data mining benchmarks: a survey and empirical demonstration.
\newblock In {\em Proceedings of the eighth ACM SIGKDD international conference on Knowledge discovery and data mining}, pages 102--111, 2002.

\bibitem{keogh2005hot}
E.~Keogh, J.~Lin, and A.~Fu.
\newblock Hot sax: Efficiently finding the most unusual time series subsequence.
\newblock In {\em Fifth IEEE International Conference on Data Mining (ICDM'05)}, pages 8--pp. Ieee, 2005.

\bibitem{KHARITONOV20221288}
A.~Kharitonov, A.~Nahhas, M.~Pohl, and K.~Turowski.
\newblock Comparative analysis of machine learning models for anomaly detection in manufacturing.
\newblock {\em Procedia Computer Science}, 200:1288--1297, 2022.
\newblock 3rd International Conference on Industry 4.0 and Smart Manufacturing.

\bibitem{liu2022ai}
C.~Liu, W.~Tian, and C.~Kan.
\newblock When ai meets additive manufacturing: Challenges and emerging opportunities for human-centered products development.
\newblock {\em Journal of Manufacturing Systems}, 64:648--656, 2022.

\bibitem{LIU2022648}
C.~Liu, W.~Tian, and C.~Kan.
\newblock When ai meets additive manufacturing: Challenges and emerging opportunities for human-centered products development.
\newblock {\em Journal of Manufacturing Systems}, 64:648--656, 2022.

\bibitem{liu2008isolation}
F.~T. Liu, K.~M. Ting, and Z.-H. Zhou.
\newblock Isolation forest.
\newblock In {\em 2008 eighth ieee international conference on data mining}, pages 413--422. IEEE, 2008.

\bibitem{nasrin2024application}
T.~Nasrin, F.~Pourkamali-Anaraki, and A.~M. Peterson.
\newblock Application of machine learning in polymer additive manufacturing: A review.
\newblock {\em Journal of Polymer Science}, 62(12):2639--2669, 2024.

\bibitem{rao2015online}
P.~K. Rao, J.~Liu, D.~Roberson, Z.~Kong, and C.~Williams.
\newblock Online real-time quality monitoring in additive manufacturing processes using heterogeneous sensors.
\newblock {\em Journal of Manufacturing Science and Engineering}, 137(6):061007, 2015.

\bibitem{reisch2020distance}
R.~Reisch, T.~Hauser, B.~Lutz, M.~Pantano, T.~Kamps, and A.~Knoll.
\newblock Distance-based multivariate anomaly detection in wire arc additive manufacturing.
\newblock In {\em 2020 19th IEEE international conference on machine learning and applications (ICMLA)}, pages 659--664. IEEE, 2020.

\bibitem{ryck2020change}
T.~D. Ryck, M.~D. Vos, and A.~Bertrand.
\newblock Change point detection in time series data using autoencoders with a time-invariant representation, 2020.

\bibitem{shi2023lstm}
Z.~Shi, A.~A. Mamun, C.~Kan, W.~Tian, and C.~Liu.
\newblock An lstm-autoencoder based online side channel monitoring approach for cyber-physical attack detection in additive manufacturing.
\newblock {\em Journal of Intelligent Manufacturing}, pages 1--17.

\bibitem{simplify3dGcode}
{Simplify3D}.
\newblock 3d printing g-code tutorial, n.d.
\newblock [Online; accessed 1-Mar-2025].

\bibitem{wu2022online}
Q.~Wu, N.~Xie, S.~Zheng, and A.~Bernard.
\newblock Online order scheduling of multi 3d printing tasks based on the additive manufacturing cloud platform.
\newblock {\em Journal of Manufacturing Systems}, 63:23--34, 2022.

\bibitem{yeh2016matrix}
C.-C.~M. Yeh, Y.~Zhu, L.~Ulanova, N.~Begum, Y.~Ding, H.~A. Dau, D.~F. Silva, A.~Mueen, and E.~Keogh.
\newblock Matrix profile i: all pairs similarity joins for time series: a unifying view that includes motifs, discords and shapelets.
\newblock In {\em 2016 IEEE 16th international conference on data mining (ICDM)}, pages 1317--1322. Ieee, 2016.

\bibitem{zhang2020semantic}
L.~Zhang, Y.~Gao, and J.~Lin.
\newblock Semantic discord: Finding unusual local patterns for time series.
\newblock In {\em Proceedings of the 2020 SIAM International Conference on Data Mining}, pages 136--144. SIAM, 2020.

\bibitem{Shi2022LSTM}
C.~K. W. T. C.~L. Zhangyue~Shi, Abdullah Al~Mamun.
\newblock An lstm-autoencoder based online side channel monitoring approach for cyber-physical attack detection in additive manufacturing.
\newblock In {\em Journal of Intelligent Manufacturing (2023)}, pages 1815--1831. Ieee, 2022.

\bibitem{zhouCASE2021}
H.~Zhou, C.~Liu, W.~Tian, and C.~Kan.
\newblock Echo state network learning for the detection of cyber attacks in additive manufacturing.
\newblock In {\em 2021 IEEE 17th International Conference on Automation Science and Engineering (CASE)}, pages 177--182, 2021.

\bibitem{zhu2016matrix}
Y.~Zhu, Z.~Zimmerman, N.~S. Senobari, C.-C.~M. Yeh, G.~Funning, A.~Mueen, P.~Brisk, and E.~Keogh.
\newblock Matrix profile ii: Exploiting a novel algorithm and gpus to break the one hundred million barrier for time series motifs and joins.
\newblock In {\em 2016 IEEE 16th international conference on data mining (ICDM)}, pages 739--748. Ieee, 2016.

\end{thebibliography}
\end{document}